\documentclass[twocolumn]{svjour3} 
\smartqed
\usepackage{graphicx}
\usepackage{subfigure}
\usepackage{booktabs}
\usepackage{multirow}
\usepackage{amsmath,amssymb}
\usepackage{color}
\usepackage{natbib}
\usepackage[colorlinks, linkcolor=red, anchorcolor=red, citecolor=blue]{hyperref}
\usepackage{pifont}
\usepackage{float}
\usepackage[table]{xcolor}

\begin{document}
\begin{sloppypar}


\title{Lightweight High-Speed Photography Built on Coded Exposure and Implicit Neural Representation of Videos}


\author{Zhihong~Zhang$^{1}$ \and Runzhao~Yang$^{1}$ \and Jinli~Suo$^{1,2,3}$ \and Yuxiao~Cheng$^{1}$ \and Qionghai~Dai$^{1,2}$}
\institute{
Zhihong Zhang \at zhangzh19@mails.tsinghua.edu.cn
\and
Runzhao Yang \at yangrz20@mails.tsinghua.edu.cn
\and
Jinli Suo \at jlsuo@tsinghua.edu.cn
\and
Yuxiao Cheng \at cyx22@mails.tsinghua.edu.cn
\and
Qionghai Dai \at qhdai@tsinghua.edu.cn
\and
$^1$ Department of Automation, Tsinghua University, Beijing, 100084, China.\at 
$^2$ Institute for Brain and Cognitive Sciences, Tsinghua University, Beijing, 100084, China\at
$^3$ Shanghai Artificial Intelligence Laboratory, Shanghai, 200030, China. 
}

\date{Received: date / Accepted: date}

\maketitle

\begin{abstract}
The demand for compact cameras capable of recording high-speed scenes with high resolution is steadily increasing. However, achieving such capabilities often entails high bandwidth requirements, resulting in bulky, heavy systems unsuitable for low-capacity platforms. To address this challenge, leveraging a coded exposure setup to encode a frame sequence into a blurry snapshot and subsequently retrieve the latent sharp video presents a lightweight solution. Nevertheless, restoring motion from blur remains a formidable challenge due to the inherent ill-posedness of motion blur decomposition, the intrinsic ambiguity in motion direction, and the diverse motions present in natural videos.  
In this study, we propose a novel approach to address these challenges by combining the classical coded exposure imaging technique with the emerging implicit neural representation for videos. We strategically embed motion direction cues into the blurry image during the imaging process. Additionally, we develop a novel implicit neural representation based blur decomposition network to sequentially extract the latent video frames from the blurry image, leveraging the embedded motion direction cues.
To validate the effectiveness and efficiency of our proposed framework, we conduct extensive experiments using benchmark datasets and real-captured blurry images. The results demonstrate that our approach significantly outperforms existing methods in terms of both quality and flexibility. The code for our work is available at \href{https://github.com/zhihongz/BDINR}{https://github.com/zhihongz/BDINR}
\end{abstract}



\section{Introduction}
Mobile platforms equipped with compact high-speed cameras are of wide applications. On the one hand, with the rapid development of the Internet, it has been a popularity for people to record their daily lives by taking photos or videos and share them with others on social media. 
Although current smartphones and digital cameras have shown excellent imaging quality in most scenarios, they struggle to capture the details of fast-moving objects and suffer from blur artifacts due to the limited frame rate \citep{rozumnyi2021DeFMODeblurring,li2022HighSpeedLargeScale}. This problem becomes even severe in low-light conditions where a longer exposure duration is required to accumulate enough photons for a better signal-to-noise ratio (SNR) \citep{li2022lowlight,sanghvi2022PhotonLimitedBlind}. 
On the other hand, acquiring high-quality videos at high speed is also one of the significant demands of photography in industry, agriculture, military, and other fields. The lightweight design applicable for low-capacity platforms is especially important, and holds great potential in a wide range of applications, such as vision navigation of self-driving cars, robots, and drones. 
Similar to the smartphones, limited load capacity and computing resources impose big challenges on the imaging technologies, and are calling for large efforts in this direction even after decades of studies.


To improve the ability of imaging systems to capture transient moments, a vast body of work in imaging sensors, computer vision, computational photography, and related fields has emerged over the last few decades. From the hardware perspective, it has come to a bottleneck to improve the overall throughput of digital cameras due to the limited on-chip memory and readout speed of imaging sensors. This constraint leads to an intrinsic trade-off between the spatial resolution and temporal resolution for video acquisition. Fortunately, the flourishing development of computer vision and deep learning in recent years has shed new light on circumventing this trade-off by exploring data-driven prior of natural images in post-processing algorithms.

On the algorithm side, various lines of works including motion deblurring, video interpolation, and blur decomposition have been proposed to remove the blur artifacts or recover the motion dynamics from the images or videos captured by low-speed cameras. Specifically, motion deblurring aims to remove the blur artifacts and restore the sharp details in the blurry video, but it doesn't improve the frame rate of the video after processing \citep{zhang2022DeepImage,rota2023VideoRestoration}. By contrast, video interpolation takes low-frame-rate videos as input and generates their high-frame-rate counterparts via temporal interpolation \citep{parihar2022ComprehensiveSurvey,dong2023VideoFrame}. Since most low-frame-rate videos exhibit some degrees of blur that cannot be eliminated through simple inter-frame interpolation, video interpolation algorithms typically require additional designs to improve the sharpness of the output video \citep{zuckerman2020ScalesDimensions}. 

Blur decomposition is another line of algorithms aimed at reversing single blurry images to sharp dynamic video clips thus improving the frame rate of the processed video \citep{jin2018LearningExtract,purohit2019BringingAlive,argaw2021RestorationVideo,li2022HighSpeedLargeScale,zhong2022AnimationBlur}. Despite its potential, blur decomposition is comparatively intricate and has garnered relatively less attention than motion deblurring or video interpolation. The complexity of this task stems from two primary challenges. Firstly, extracting video sequences from single blurry images is inherently ill-posed, with a significantly higher level of ill-posedness compared to image deblurring or video interpolation. Secondly, motion-blurred images arise from the accumulation of instantaneous frames during sensor exposure. This accumulation process disrupts the temporal order of the individual frames, thus leading to motion direction ambiguity problem in blur decomposition \citep{jin2018LearningExtract,purohit2019BringingAlive}. 
Motion direction ambiguity constitutes an inherent issue in blur decomposition that defies complete resolution through post-processing algorithms. This is because the motion trajectories with opposite directions result in identical blur kernels, leading to indistinguishable blurry images. Consequently, when presented with a motion-blurred image, discerning the precise motion direction becomes an insurmountable task. In real-world scenarios, objects in motion can move in various directions or their opposites. As a result, while blur decomposition networks may successfully reconstruct a coherent sharp video through learned priors, the inferred motion direction may not align with the actual scenario.
Furthermore, motion direction ambiguity is independent for different objects within the blurry image, giving rise to multiple combinations of possible motion directions for different objects in the solution space. This makes the complexity of blur decomposition exponentially expand as the number of dynamic objects increases.
A detailed analysis of this issue is presented in Sec.~\ref{sec:ce_embed}

\begin{figure*}[!ht]
\begin{center}
  \includegraphics[width=1\linewidth]{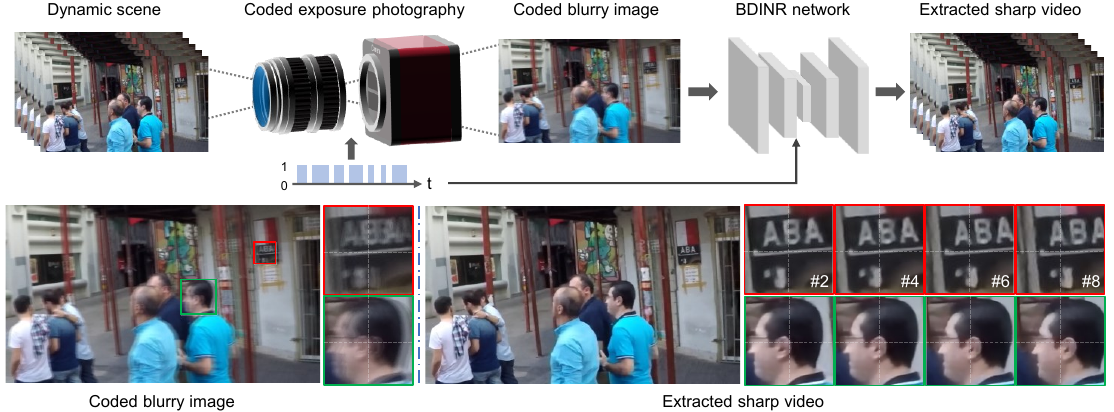}
\end{center}
\caption{\textbf{The overall schematic and demo results of the proposed blur decomposition framework.} On the imaging side, coded exposure photography is employed to embed motion direction cues into the captured coded blurry image. It also facilitates the information preservation of the blurry image across all frequencies. On the algorithm side, a video INR based self-recursive blur decomposition network (BDINR) is developed to extract the latent video sequence collapsed in the coded blurry image by exploiting the embedded motion direction cues.} 
\label{fig:flowchart}
\end{figure*}

To cope with these issues, some works simplify the blur decomposition problem with extra assumptions about the number and motion types of the animated objects \citep{rozumnyi2021DeFMODeblurring,li2022HighSpeedLargeScale}. These assumptions narrow down the solution space thus making the problem easier to solve, but they also limit corresponding methods' applications and performance in practical scenarios. Other approaches rely on introducing additional information like manually annotated motion directions \citep{zhong2022AnimationBlur} or event sequences from event cameras \citep{pan2019BringingBlurry,lin2020LearningEventDriven} as supplementary clues to finish the video extraction task, but such information may be inaccessible in conventional cases. 

Bearing the limitations of pure hardware-based and algorithm-based approaches in mind, we propose a novel computational photography based blur decomposition framework by jointly designing the imaging system and the post-processing algorithm as shown in Fig.~\ref{fig:flowchart}. Specifically, on the imaging side, we conduct a comprehensive analysis of the blur formation process of different imaging paradigms and figure out their relationship with the motion ambiguity issue in the blur decomposition problem. We find that the classical coded exposure photography technique not only facilitates information preservation in motion-blurred images, but also has potential in implicitly embedding the motion direction cues into the captured coded blurry image to cope with the motion ambiguity challenge. On the algorithm side, inspired by recent advances in implicit neural representation (INR), we propose to represent the latent sharp video sequence encoded in the coded blurry image with a learnable video INR and incorporate it into a self-recursive neural network to sequentially extract the latent frames by exploiting the embedded motion direction cues. Benefiting from the efficient representation ability of video INR, the designed network encompasses notable properties such as small size, superior blur decomposition performance, and exceptional flexibility in practical applications.


In a nutshell, we propose a novel blur decomposition framework by combining coded exposure photography and a video INR based self-recursive neural network. The main contributions of this work can be summarized as follows:
\begin{itemize}
\item[-] We delve into the motion direction ambiguity issue in blur decomposition problem and propose to introduce coded exposure photography technique for implicit motion direction embedding to deal with this issue. 
\item[-] We develop a video INR based self-recursive neural network to sequentially decompose the latent sharp video frames from a single coded blurry image by exploiting the embedded motion direction cues. The network features small size, superior performance, and high flexibility.
\item[-] We conduct comprehensive experiments on both simulated data and real data to validate the effectiveness and efficiency of the proposed framework. The results demonstrate that the proposed framework significantly outperforms existing approaches.
\end{itemize}


\section{Related Work}
\label{sec:related_work}
In this section, we firstly review the learning-based approaches for blur decomposition and outline the open challenges in this field. Then, we provide a brief overview of coded exposure photography and video INR, along with their recent applications relevant to this research.

\subsection{Blur Decomposition}
\label{sec:bd}

Blur decomposition is an emerging but promising field aiming to extract a video sequence from a single motion-blurred image. As mentioned above, the difficulty mainly lies in the high ill-posedness and motion direction ambiguity caused by the accumulation of instant frames during the exposure. with the aid of the powerful representation ability of deep neural networks (DNNs), some learning-based algorithms have been proposed to solve the blur decomposition problem in recent years \citep{jin2018LearningExtract,purohit2019BringingAlive,argaw2021RestorationVideo,zhong2022AnimationBlur,pan2019BringingBlurry,yosef2023VideoReconstruction}. 

In 2018, Jin et al. initially formulated the problem of blur decomposition and gave a comprehensive analysis on the ambiguity of temporal ordering and motion directions which makes the problem challenging \citep{jin2018LearningExtract}. They also designed the first learning-based method and introduced a temporal-order invariant loss as the regularizer to sequentially extract pairs of frames to generate a video from a single motion-blurred image. Later, Purohit et al. proposed a two-stage deep convolutional architecture for blur decomposition \citep{purohit2019BringingAlive}. They firstly trained a recurrent video auto-encoder in a self-supervised manner to learn motion representation from sharp videos. Then they replaced the video encoder with a blurred image encoder, and optimized the newly formed auto-encoder for video extraction from a blurry image. Argaw et al. also adapted an encoder-decoder structure but introduced the spatial transformer as basic modules in their end-to-end blur decomposition network \citep{argaw2021RestorationVideo}. They further delicately designed their loss functions and introduced extra regularizers with complementary properties to stabilize the training. 

It is worth noting that, even though these methods could attain plausible results in some simple cases by incorporating deep neural networks with delicate handcrafted regularizers, none of them substantially address the ambiguity challenge, and theoretically, they cannot distinguish the forward and backward motions.
To circumvent this issue, some works attempted to explicitly introduce additional information on motion directions and achieved superior performance to previous methods. For example, Pan et al. proposed an Event-Based Double Integral (EDI) model for restoring high-frame-rate videos from a single blurry image \citep{pan2019BringingBlurry}. The model utilizes additional event data from an event camera to provide motion direction clues and outperforms prior methods. Nevertheless, its implementation suffers from increased system bulk and cost. Zhong et al. proposed to introduce supplementary motion guidance to assist the blur decomposition and designed a unified framework that supported various interfaces for motion guidance input \citep{zhong2022AnimationBlur}. However, in practical scenarios, it's difficult even impossible to manually record the motion directions of all the moving objects when taking pictures or artificially recognize the motion directions from a single blurry image afterward. 



In summary, as early attempts to solve the blur decomposition problem, existing methods generally circumvent the fundamental issues of high ill-posedness and motion direction ambiguity by designing sophisticated regularizers or introducing additional motion clues. These strategies greatly limit their applications in practice and are prone to fail in cases with multiple objects or complicated motions. In this work, by revisiting the classical coded exposure imaging technique, we tactfully embed the motion direction cues into the blurry images themselves during the imaging process. Cooperated with a specially designed blur decomposition network, the proposed method effectively puts a step forward for improving blur decomposition's performance in practical applications. 

\subsection{Coded Exposure Photography}
\label{sec:ce}

Coded exposure photography stands as a representative computational photography technique initially proposed by Raskar et al. \citep{raskar2006CodedExposure} to facilitate motion deblurring \citep{agrawal2009InvertibleMotion,mccloskey2010VelocityDependentShutter,harshavardhan2013FlutterShutter,zhang2023DeepCoded}. Unlike conventional photography, where the camera’s shutter remains open throughout the entire exposure duration, the coded exposure technique intermittently opens and closes the camera’s shutter based on a predetermined binary sequence during the exposure period. This method allows us to tailor the blur kernel of motion-blurred images to have no zero points and relatively flat magnitude across the entire frequency spectrum, thereby preserving information from all frequencies in the coded blurry image. In contrast, images of moving targets captured under conventional exposure result in box blur, characterized by a sinc-function form frequency spectrum. This type of blur acts as a low-pass filter and exhibits periodic zero points in the frequency domain, leading to information loss at corresponding frequency points.

While previous studies related to coded exposure photography predominantly focus on optimizing encoding sequences or developing corresponding algorithms to improve motion deblurring performance \citep{agrawal2009OptimalSingle,agrawal2009CodedExposure,jeon2015ComplementarySets,jeon2017GeneratingFluttering,cui2021EffectiveCoded,zhang2023DeepCoded}, it is important to recognize that coded exposure photography, acting as a front-end physical approach to enhance image information preservation, also harbors the potential to facilitate blur decomposition algorithms. Therefore, in this study, we capitalize on the advantageous properties of coded exposure photography in preserving high-frequency information and extend its applicability to addressing the motion ambiguity issue in blur decomposition by introducing the "asymmetry" constraint to the exposure encoding sequence design. A comprehensive explanation will be provided in Sec.~\ref{sec:ce_embed}.

It is worth noting that there is another line of works called pixel-wise coded exposure, which is also known as coded aperture compressive temporal imaging (CACTI) or snapshot compressive imaging (SCI) \citep{hitomi2011VideoSingle,llull2013CodedAperture,dengyuliu2014EfficientSpaceTime,deng2021SinusoidalSampling,zhang2021TenmegapixelSnapshot}. Different from the aforementioned coded exposure photography which flutters the camera shutter globally during an exposure, the pixel-wise coded exposure technique enables pixel-level exposure control and can be used for recovering video sequences from blurry images, similar to blur decomposition. However, pixel-wise coded exposure imaging generally requires an additional spatial light modulator and corresponding relay optics, which increase the complexity and cost of the system. Moreover, on account of the strict demand for pixel-level alignment, it's also sensitive to external disturbance and requires tedious calibration before data acquisition. In contrast, the global coded exposure can be directly realized using any commercial camera that supports IEEE DCAM Trigger Mode 5 \citep{agrawal2009CodedExposure, jeon2015ComplementarySets, mccloskey2012DesignEstimation}. It doesn't require fussy calibration and is robust to diverse environmental conditions in practical application.

In this study, we present an in-depth analysis of the relation between the blurry image formation process and the motion direction ambiguity issue under different exposure conditions (see Sec.~\ref{sec:ce_embed}). The analysis demonstrates that employing the cost-efficient and easy-to-implement conventional coded exposure technique (i.e. flutter shutter) can effectively embed the motion direction cues into the coded blurry image, thus providing implicit motion guidance to facilitate the blur decomposition task.

\subsection{Implicit Neural Representation for Videos}
\label{INRV}

Implicit neural representation (INR) provides a novel approach for parameterizing various signals including images, videos, 3D scenes, etc. \citep{mildenhall2020NeRFRepresenting,chen2021NervNeural,karras2021AliasfreeGenerative,yang2022SCISpectrum}. The fundamental concept behind INR is to model a signal as a function that can be approximated using a neural network. This neural network implicitly encodes the signal's values in its architecture and parameters during training/fitting, and these values can be retrieved through corresponding coordinates afterward. According to the universal approximation theorem of neural networks, an INR implemented with Multi-Layer Perceptron (MLP) is capable of fitting highly intricate functions by utilizing an adequate number of parameters \citep{pinkus1999ApproximationTheory}. 

Neural representation for videos is a special line of INR-based approaches focusing on parameterizing videos with neural networks. Currently, image-wise video INR has become dominant in this field \citep{chen2021NervNeural,li2022ENeRVExpedite,chen2023HNeRVHybrid}. Chen et al. proposed the first image-wise video INR approach called NeRV and demonstrated its applications in video compression and video denoising \citep{chen2021NervNeural}. Implemented with an MLP + ConvNets architecture, NeRV takes frame indexes as input and directly outputs corresponding video frames. Compared with conventional pixel-wise representation methods that map pixel coordinates to corresponding RGB values, NeRV shows significant advantages in sampling speed and representation quality. Later, Li et al. proposed E-NeRV by upgrading NeRV's redundant network structure and disentangling the spatial-temporal context in the image-wise INR \citep{li2022ENeRVExpedite}. ENeRV significantly expedites NeRV and achieves an $8 \times$ faster convergence speed. Most recently, Chen et al. further proposed HNeRV by optimizing NeRV's network architecture with a novel HNeRV block and substituting learnable and content-adaptive frame index embeddings for previous fixed and content-agnostic ones used in NeRV and ENeRV \citep{chen2023HNeRVHybrid}. Compared with NeRV and ENeRV, HNeRV demonstrates superior performance on both reconstruction quality and convergence speed.

As a novel and efficient scheme for visual signal representation, INR has also been utilized in developing innovative algorithms to address video-related computer vision challenges. For example, Shangguan et al. introduced INR into the problem of temporal video interpolation and proposed a new learning-based algorithm named CURE \citep{shangguan2022LearningCrossVideo}. 
Similarly, taking advantage of INR's continuous representation ability, Chen et al. showed video INR's application in continuous space-time super-resolution and significantly outperformed prior approaches \citep{chen2022VideoINRLearning}. Mai et al. designed a motion-adjustable video INR by mapping the temporal index of a video frame to the phase-shift information of the frame's Fourier-based position encoding. By manipulating the phase shift, the proposed method can realize motion magnification, motion smoothing, and video interpolation \citep{mai2022MotionadjustableNeural}. 

In this work, we incorporate an image-based video INR into a new blur decomposition network called BDINR to implicitly represent the latent sharp video hidden behind the corresponding blurry image. BDINR conditions the video INR on the physical blurring model and introduces a self-recursive architecture to leverage the inter-frame correlation among the video frames. By training BDINR with blurry image and corresponding sharp video pairs, it endows the embedded video INR with the ability to learn a prior across multiple videos.

\begin{figure*}
\begin{center}
  \includegraphics[width=0.9\linewidth]{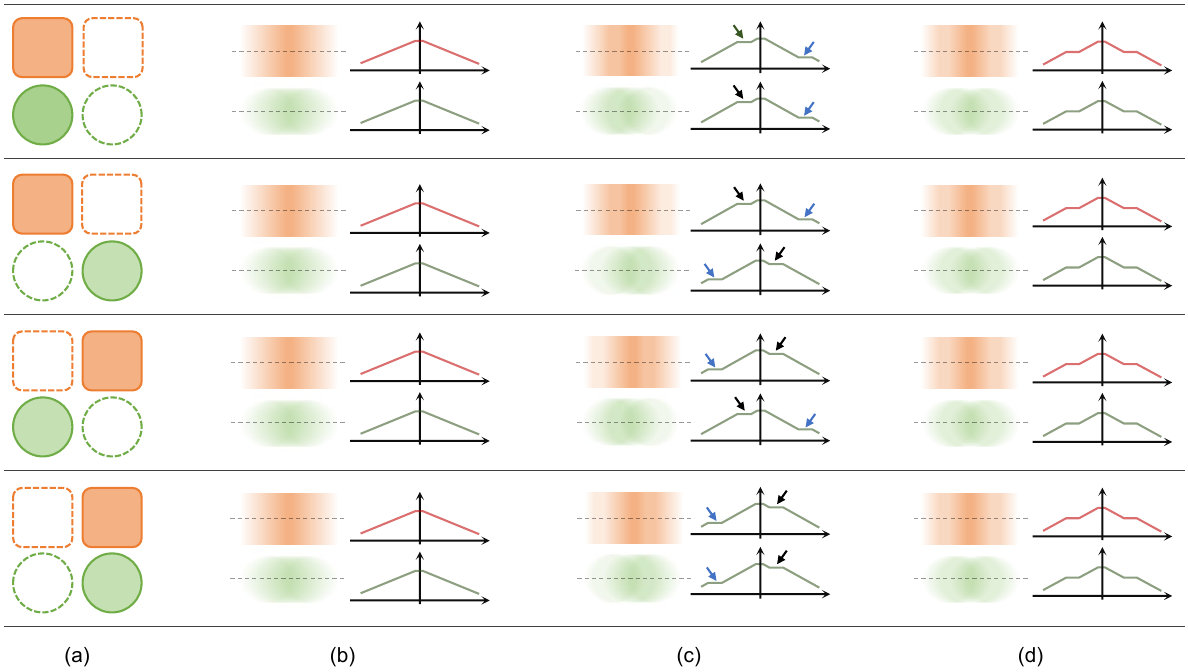}
\end{center}
\vspace{-2mm}
\caption{\textbf{Motion ambiguity in blur decomposition and motion direction embedding via coded exposure.} In this toy example, we use two horizontally translating objects, i.e., the orange cube and the green ball, for a demonstration. (a) shows four possible motion scenarios of these two objects. They translate from current positions to the dashed boxes/circles for the same distance. (b), (c), and (d) show the resulting blurry images captured under conventional exposure (`11111'), coded exposure with an asymmetric encoding sequence (`11101'), and coded exposure with a symmetric encoding sequence (`11011'), respectively. The center-line intensity profiles of the blurry images are also plotted on their right side. (c) demonstrates that employing coded exposure with an asymmetric encoding sequence will result in asymmetric blurry profiles, from which the moving direction can be retrieved (i.e. from the black arrow towards the blue arrow). Conversely, the other two cases shown in (b) and (d) will result in the same blurry images for different combinations of motion directions, thus causing the motion direction ambiguity issue in blur decomposition.}
\label{fig:motion_embed}
\end{figure*}

\section{The Proposed Method}
\label{sec:our_method}

To deal with the challenges of high ill-posedness and motion direction ambiguity, we present a novel blur decomposition framework by incorporating coded exposure photography and implicit neural representation. Specifically, on the imaging side, we take advantage of coded exposure photography's superior information-preservation ability and further employ it as an efficient tool for motion direction embedding. On the algorithm side, we represent the latent video sequence encoded in the coded blurry image with a video INR, and develop a novel self-recursive neural network to sequentially retrieve the latent frames with the aid of data-driven prior and embedded motion direction cues.

\subsection{Coded Exposure based Motion Direction Embedding} 
\label{sec:ce_embed}
\noindent\textbf{Mathematical formulation.\quad} 
The imaging process of digital cameras 
can be physically modeled as the integration of scene radiance on the sensor during an exposure elapse, and generally long-exposure photography of a dynamic scene will result in a blurry image. 
In coded exposure photography with a binary encoding sequence, the entire exposure duration is divided into several isometric segments, and each segment corresponds to a bit in the encoding sequence that controls the flutter's open/close state. 
Specifically, `1' triggers the open state with scene radiance accumulated on the sensor, 
while `0' triggers the close state blocking the incoming light. 
Mathematically, the formation of blurry measurement in coded exposure photography can be formulated as 
\begin{equation}
\label{eq: ce_model1}
\mathbf{B} = \int_{t=0}^{T}\mathbf{I}(t)\mathbf{c}(t)dt,
\end{equation}
where $\mathbf{B}$ denotes the coded blurry snapshot, $T$ is the total exposure duration, $\mathbf{I}(t)$ and 
$\mathbf{c}(t)$ are the scene intensity and shutter's trigger signal at time $t$, respectively. Note that, for concision and clarity, we omit the camera's response function and post-processing steps like digital gain and gamma transformation, which can be calibrated and compensated beforehand in practical applications. Eq.~\eqref{eq: ce_model1} can be further discretized into
\begin{equation}
\label{eq: ce_model2}
\mathbf{B} = \sum_{n=1}^{N}\mathbf{I}_n\mathbf{c}_n,
\end{equation}
where $N$ represents the length of the exposure encoding sequence, i.e. the number of brief exposure segments.

\vspace{1mm}
\noindent\textbf{Motion direction embedding.\quad} 
To depict the underlying principles of motion ambiguity in blur decomposition and embedding of motion direction via coded exposure, we illustrate a simple toy example in Fig.~\ref{fig:motion_embed}. This toy example demonstrates two objects (an orange cube and a green ball) shifting horizontally from current positions to the dashed boxes/circles for the same distance, forming four types of motion combinations as shown in column (a). Under conventional exposure shown in column (b), four acquired blurry images and their center-line intensity profiles plotted on the right are exactly the same. In other words, given the captured blurry image, it is impossible to determine the actual motion directions of these two objects, which is referred to as motion direction ambiguity in blur decomposition. In real scenarios involving more dynamic objects and complex motion trajectories, the ambiguity aggravates and thus the blur decomposition task gets more challenging. 

However, by introducing coded exposure imaging with specially designed encoding sequences, this issue can be subtly mitigated. Recalling that coded exposure turns the smearing blur into fringes i.e., discontinuous blur profile (please refer to the right parts of column (c) and (d)), with the intensity variation along the profile corresponding to the coding sequence.  
Therefore, when a symmetric encoding sequence (`11011') is used (column (d)), the resulting blurry images are still the same for these four cases. But for an asymmetric encoding sequence (`11101') shown in column (c), things turn around --- the intensity profiles become asymmetric accordingly, and thus the blurry images from four different scenarios are distinguishable. In this case, the asymmetrically located `0's in the encoding sequence act like a unique `timestamp', which together with the `milestone', i.e. the asymmetrical discontinuous blur profile, can help retrieve the motion directions successfully. 

Based on above analysis, we select the coded-exposure encoding sequence according to the following criteria:
\begin{itemize}
\item[-] The encoding sequence ought to exhibit asymmetry to effectively fulfill its role in embedding motion direction, thereby mitigating motion direction ambiguity during blur decomposition.
\item[-] The frequency spectrum of the encoding sequence should ideally have a large minimum value and a low variance, which facilitates enhanced preservation of information across diverse frequencies within the coded blurred image. \citep{raskar2006CodedExposure}.
\end{itemize}

In practice, we begin by utilizing Raskar's method \citep{raskar2006CodedExposure} to identify several encoding sequence candidates with favorable spectrum properties. Subsequently, from this pool, we select an asymmetric sequence for experimentation. For a comprehensive understanding of the encoding sequence's impact on the final blur decomposition performance, please refer to Section~\ref{sec:analysis}, where we provide detailed insights.

It is noteworthy that while coded exposure photography aids in eliminating motion direction ambiguity and minimizing high-frequency information loss in blurry images, blur decomposition remains an intrinsic ill-posed problem characterized by spatial-temporal information aliasing. To address this challenge, we propose a novel learnable video INR-based blur decomposition network in the subsequent subsection. This network leverages the exceptional capacity of deep neural networks in handling ill-posed problems through data-driven priors to reconstruct the sharp underlying video.

\subsection{Video INR based Self-recursive Blur Decomposition} 

\begin{figure*}[h]
\begin{center}
  \includegraphics[width=1\linewidth]{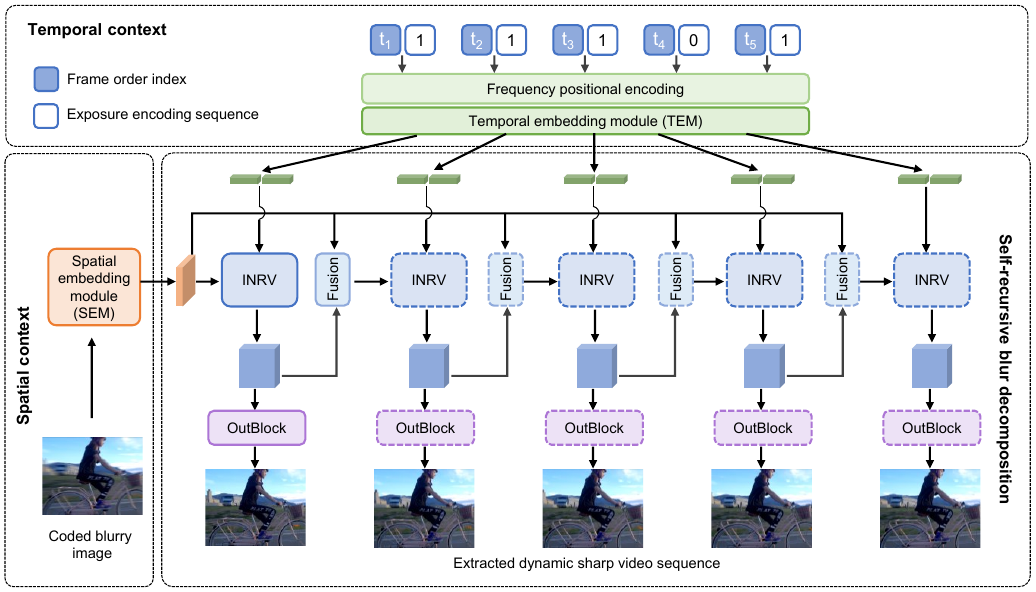}
\end{center}
\caption{\textbf{The overall flowchart of the proposed video INR based self-recursive blur decomposition network (BDINR).} The temporal embedding module (TEM) fuses the frame order index and corresponding exposure-encoding sequence to generate the temporal context embedding. The spatial embedding module (SEM) maps the coded blurry image into a continuous feature space to serve as the spatial context embedding. These embeddings are then input to the video INR module (INRV) for latent frame extraction in a self-recursive manner.} 
\label{fig:cebd}
\end{figure*}

\begin{figure*}[h]
\begin{center}
  \includegraphics[width=0.9\linewidth]{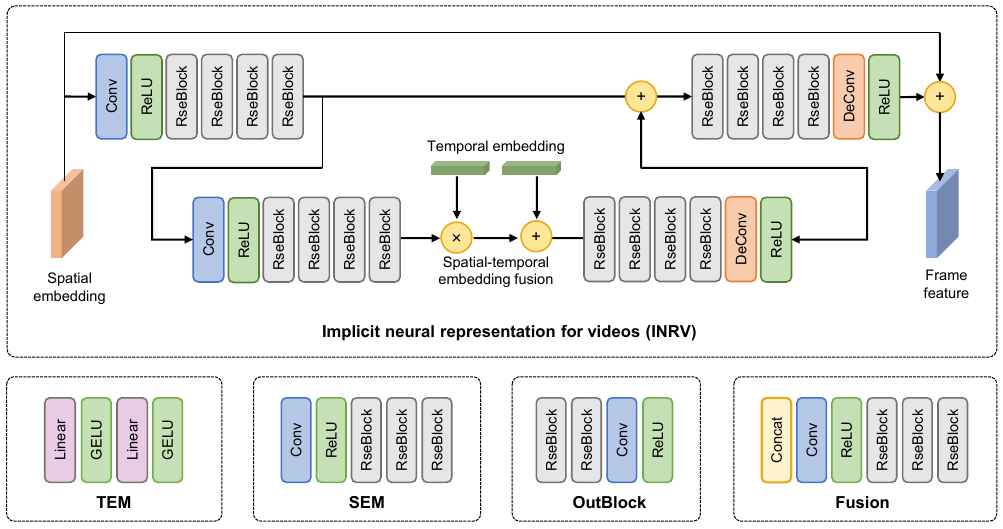}
\end{center}
 \vspace{-2mm}
\caption{\textbf{The specific network structure of different modules involved in BDINR.} INRV comprises a two-level encoder-decoder architecture to fuse the spatial and temporal embeddings. TEM is implemented with a two-layer perceptron. SEM and the rest of the modules are mainly composed of convolutional layers and residual blocks.} 
\label{fig:cebd_detail}
\end{figure*}

To efficiently utilize the motion direction cues embedded in the coded blurry images, we develop a novel video INR empowered self-recursive blur decomposition network named BDINR. 
Generally, conventional video INR approaches directly map spatial-temporal coordinates to pixel values of latent frames. However, it can be challenging or even impossible to learn such a video INR merely from the given blurry image in light of the highly ill-posed nature of the blur decomposition problem. 
Therefore, we disentangle the temporal and spatial context information with the temporal embedding module (TEM) and the spatial embedding module (SEM) in BDINR and further develop a learnable image-based video INR module (INRV) to introduce data-driven prior for mitigating the ill-posedness through supervised training.

The overall flowchart of BDINR is illustrated in Fig.~\ref{fig:cebd}. As shown in the figure, TEM takes as input the frame order index and corresponding exposure-modulation sequence, encoded as frequency position encoding, to generate the temporal context embedding. Similarly, SEM takes the coded blurry image as input to generate the spatial context embedding. Finally, INRV fuses these temporal and spatial context embeddings to retrieve the corresponding latent video sequence in a self-recursive manner. The detailed design of each part is depicted in Fig.~\ref{fig:cebd_detail} and described below.

\vspace{1mm}
\noindent\textbf{TEM and SEM.\quad}
Prior to TEM, we employ the frequency position encoding strategy to map the frame index into a high-dimensional embedding space, which enhances the network's capacity in fitting data with high-frequency variations\citep{tancik2020FourierFeatures,mildenhall2020NeRFRepresenting,li2022ENeRVExpedite}. 
Meanwhile, we also incorporate the binary exposure code into the phase of the position encoding, which helps to identify the occurrence of the corresponding frame in the coded blurry image. Mathematically, the position encoding function $\gamma(\cdot)$ can be formulated as
\begin{gather}
\label{eq: pe}
\gamma(\mathbf{t}_i, \mathbf{c}_i) =  \left[\sin(b^0\pi \mathbf{t}_i + \widehat{\mathbf{c}}_i\pi), \cos(b^0\pi \mathbf{t}_i + \widehat{\mathbf{c}}_i\pi), \dots, \right. \notag \\
            \quad\quad\quad\quad~~ \left.\sin(b^{l-1}\pi \mathbf{t}_i + \widehat{\mathbf{c}}_i\pi), \cos(b^{l-1}\pi \mathbf{t}_i + \widehat{\mathbf{c}}_i\pi)\right], \\
\widehat{\mathbf{c}}_i = 1 - \mathbf{c}_i,
\end{gather}
where $\mathbf{t}_i$ and $\mathbf{c}_i$ denote the normalized frame index and the binary code corresponding to the $i_{th}$ latent frame, respectively. $b$ and $l$ are two hyper-parameters, which are empirically set to 1.25 and 80 in our experiments.

TEM is implemented with a two-layer perceptron, fusing the frequency position encoding of the frame index to efficiently model the temporal correlation. During video extraction, the temporal context embedding generated by TEM helps to provide temporal information for locating the desired latent frame in the INRV. Moreover, it also serves as an index to retrieve the motion direction cues hidden in the blurry image. 

SEM is composed of a convolutional layer and three residual blocks, embedding the discrete input coded blurry image into a continuous spatial feature space. The embedded feature serves as an informative spatial context reference for the latent frames encoded by the INRV. In this manner, SEM not only reduces the INRV's representation difficulty but also enables the INRV to learn a prior through supervised training from massive pairs of blurry images and the ground-truth latent video. 

\vspace{1mm}
\noindent\textbf{INRV module and self-recursive strategy.\quad}
We represent the latent video sequence underlying the coded blurry image with an image-based video INR due to its powerful representation ability of continuous signals. The video INR module (INRV) is implemented with an encoder-decoder architecture. The encoder firstly takes the spatial context embedding as input and further maps it into a deeper high-dimensional feature space. Then it will be fused with the temporal embedding through a simple linear transformation to combine the spatial-temporal information and eliminate the motion ambiguity. The fused embedding serves as a unique index and will finally be input to the decoder for the retrieval of the corresponding latent frame.

To make adequate use of the temporal correlation among the latent video frames and reduce the model size, we further employ a self-recursive strategy during the frame extraction process. Specifically, starting from the second frame, we fuse the spatial context embedding generated from the coded blurry image with the output feature of the previous extracted frame to serve as a new spatial context embedding. In this manner, the retrieval of each frame can receive additional guidance from its previous frames. Note that, in Fig.~\ref{fig:cebd}, we unfold the frame extraction process into multiple steps for clear demonstration, but there is essentially only one set of modules including INRV, Fusion, and OutBlock.   

Moreover, unlike conventional video INR approaches that fit specific signals with network parameters and form a one-to-one mapping, our proposed BDINR efficiently exploits the powerful representation ability of deep networks and the redundancy of natural images to learn a one-to-more mapping video INR conditioned on the physical model of coded exposure photography. In this sense, the learned video INR can be regarded as a meta-video dictionary, from which specific video sequences can be retrieved with the coded blurry image and corresponding exposure encoding sequence as indexes.

\vspace{1mm}
\noindent\textbf{Loss function.\quad} 
We incorporate a supervised blur decomposition loss and an unsupervised reblur loss to optimize the BDINR. The blur decomposition loss penalizes large deviation of the extracted latent video sequence from the ground truth and comprises three terms, including Charbonnier loss \citep{charbonnier1994TwoDeterministic}, SSIM loss, and edge loss defined as
\begin{align}
\label{eq: bd_loss}
&\mathcal{L}_{char} = \frac{1}{P}\sum_{n=1}^N\sqrt{\|\mathbf{\hat{I}}_n - \mathbf{I}_n\|^2 + \epsilon^2},\\
&\mathcal{L}_{ssim} = \frac{1}{P}\sum_{n=1}^N\left(1-\mathcal{F}_{ssim}(\mathbf{\hat{I}}_n, \mathbf{I}_n)\right), \\
&\mathcal{L}_{edge} = \frac{1}{P}\sum_{n=1}^N\sqrt{\|\Delta\mathbf{\hat{I}}_n - \Delta\mathbf{I}_n\|^2 + \epsilon^2},
\end{align}
where $\mathbf{\hat{I}}_n$ and $\mathbf{I}_n$ are the $n_{th}$ retrieved sharp frame and corresponding ground truth; $P$ and $N$ respectively denote the number of pixels and frames in the latent video; $\epsilon$ is a small constant set to ${10^{-3}}$; $\mathcal{F}_{ssim}(\cdot)$ and $\Delta$ represent the function for SSIM calculation and the Laplacian operator. The final blur decomposition loss is calculated as the weighted summation of the above three components:
\begin{equation}
    \mathcal{L}_{bd} = \alpha_1 \mathcal{L}_{char} + \alpha_2 \mathcal{L}_{ssim} + \alpha_3 \mathcal{L}_{edge},
\end{equation}
where $\alpha_1$, $\alpha_2$, and $\alpha_3$ are empirically set to 1.0, 0.05, and 0.05, respectively.

We additionally introduce an unsupervised reblur loss \citep{chen2018Reblur2DeblurDeblurring,nah2021clean,zhang2023INFWIDEImage} to guarantee the consistency between the extracted video sequence and corresponding coded blurry input
\begin{equation}
\label{eq: loss}
\mathcal{L}_{reblur} =  \frac{N}{P}\sqrt{\|\sum_{n=1}^{N}\mathbf{\hat{I}}_n\mathbf{c}_n-\mathbf{B}\|^2 + \epsilon^2},
\end{equation}
where $\mathbf{c}$ and $\mathbf{B}$ denote the exposure encoding sequence and the coded blurry input. The final loss is defined as $\mathcal{L} = \gamma_1 \mathcal{L}_{bd} + \gamma_2 \mathcal{L}_{reblur}$ with the hyper-parameters $\gamma_1$, and $\gamma_2$ empirically being set to 1.0 and 0.2, respectively.

\section{Experiments and Discussions}
\label{sec:exp}

\subsection{Dataset, Implementation details, and Metrics}

\noindent\textbf{Dataset.\quad}
We employ the widely used high-frame-rate video dataset GoPro \citep{nah2017DeepMultiscale} in our experiments. GoPro is captured using a GOPRO4 Hero Black camera at 240 frames per second (FPS). It comprises 33 videos, consisting of approximately 35,000 frames in total. Two-thirds of the videos are used for training and the rest for testing. 
To evaluate the model's generalization ability, we additionally introduce the WAIC TSR dataset \citep{zuckerman2020ScalesDimensions} which contains 25 videos of very complex fast dynamic scenes for performance evaluation. 

During both training and evaluation, the blurry images are synthesized using the widely-used `frame-averaging' method. Specifically, we employ a sliding window approach to sample a constant number of consecutive video frames from the datasets. For conventional exposure imaging mode, the sampled frames are directly averaged to generate a blurry image. In contrast, for coded exposure imaging mode, as per Eq.~\eqref{eq: ce_model2}, the sampled frames are first weighted by the exposure encoding sequence before taking an average. In both cases, the synthesized blurry images are further normalized to [0,1] and are injected with Gaussian noise with standard deviations uniformly sampled from [0,0.01] to enhance the robustness of the trained model in practical applications. Unless explicitly stated otherwise, we set the length of the sliding window to 8 and adopt the encoding sequence for coded exposure as `11100101' in the subsequent experiments.

\vspace{1mm}
\noindent\textbf{Implementation details.\quad}
We implement the proposed network using PyTorch \citep{paszke2019PyTorchImperative} and employ the Adan optimizer \citep{xie2023AdanAdaptive}, with $\beta_1 = 0.98$, $\beta_2 = 0.92$, and $\beta_3 = 0.99$ for parameter updating. The learning rate is initialized to $5 \times 10^{-4}$ and gradually decayed to $1 \times 10^{-6}$ using the cosine annealing strategy \citep{loshchilov2017SGDRStochastic} after two rounds of warmup. 
During training, we randomly crop the input to $256 \times 256$ pixels and employ random flipping and rotation as data augmentation tricks. The model is trained for 500 epochs with a batch size of 8. All experiments are conducted on a workstation equipped with an AMD EPYC 7H12 CPU and an NVIDIA GeForce RTX 3090 GPU.

\vspace{1mm}
\noindent\textbf{Metrics.\quad} To assess the performance of different algorithms, we utilize full-reference image quality assessment metrics, including peak signal-to-noise ratio (PSNR), structural similarity index measure (SSIM) \citep{wang2004ImageQuality}, and learned perceptual image patch similarity (LPIPS) \citep{zhang2018UnreasonableEffectiveness}, in our simulation experiments. Additionally, for our real-world experiments, we employ blind image quality assessment metrics, including MUSIQ \citep{ke2021MUSIQMultiscale} and DBCNN \citep{zhang2020BlindImage}. Higher scores on PSNR, SSIM, DBCNN, and MUSIQ metrics indicate better reconstruction quality, while for LPIPS, lower scores are preferable. Furthermore, we provide insights into the model size and the number of multiply-accumulate operations (MACs) to evaluate the efficiency of the models.

\begin{table*}
    \centering
    \caption{\textbf{Quantitative performance comparison with baseline blur decomposition and blurry video interpolation algorithms on GoPro and WAIC TSR datasets.} The computational complexity, measured in terms of MACs, and the time required for extracting or interpolating a single frame are evaluated based on $256\times256$ image patches.} \textbf{Bold} and \underline{underline} highlight the best and second-best scores, respectively.
    \begin{center}
    \resizebox{1\textwidth}{!}{
    \begin{tabular}{llccccccc}
    \toprule
    & & Jin et al.& Shedligeri et al. & Animation from Blur  & Slow Motion & BIN & BDINR \\ 
    & & \cite{jin2018LearningExtract} & \cite{shedligeri2021UnifiedFramework} & \citep{zhong2022AnimationBlur} & \citep{jin2019LearningExtract} & \citep{shen2020BlurryVideo} & (Ours) \\ \midrule    
    \multirow{3}{*}{GoPro} & PSNR (dB) & 26.35  & \underline{26.94}  & 24.25  & 26.20  & 25.52  & \textbf{30.64}  \\
    & SSIM  & 0.8169 & \underline{0.8367} & 0.7785 & 0.8064 & 0.8082 & \textbf{0.9155} \\
    & LPIPS  & \underline{0.1426} & 0.1940  & 0.2804 & 0.2685 & 0.1953 & \textbf{0.1202} \\ \midrule
    \multirow{3}{*}{WAIC TSR} & PSNR (dB) & 26.47  & 26.93  & 25.83  & \underline{27.89}  & 26.82  & \textbf{28.46}  \\
    & SSIM  & 0.8468 & 0.8503 & 0.8361 & \underline{0.8615} & 0.8525 & \textbf{0.8845} \\
    & LPIPS  & 0.1762 & \underline{0.1708} & 0.2374 & 0.2001 & 0.1738 & \textbf{0.1378} \\ \midrule
    \multirow{2}{*}{Model info.} & Param. \# (M) & 18.21  & 8.60   & 32.50  & \textbf{0.85}   & \underline{1.14}   & 3.70   \\
    & MACs (G) & 211.31 & \underline{52.26}  & 224.09 & \textbf{48.83}  & 261.80 & 242.66 \\
    & Time/frame (ms) & 15.14 & \textbf{0.68}  & 4.52 & \underline{1.12}  & 8.33 &6.35 \\
    \bottomrule
    \end{tabular}}
    \end{center}
    \label{tab:simu_res}
    \vspace{-3mm}
\end{table*}

\subsection{Baseline Algorithms for Performance Comparison}
The problem of blur decomposition is a recently emerging challenge with limited available open-source algorithms.
To quantitatively demonstrate the advantageous performance of our framework, we compare against the leading methods proposed by \cite{jin2018LearningExtract}, \cite{shedligeri2021UnifiedFramework}, and \cite{zhong2022AnimationBlur}  \footnote{Methods without open-source codes \citep{purohit2019BringingAlive,zhang2020EveryMoment,argaw2021RestorationVideo} are not included in the comparison}. Considering that our framework can also be directly applied to blurry videos without modification in a frame-wise manner, we further incorporate two blurry video interpolation algorithms \citep{jin2019LearningExtract,shen2020BlurryVideo} for performance comparison. Note that these interpolation methods take as input multiple adjacent blurry frames to recover a sharp video sequence with a higher frame rate. Therefore, they suffer from much less motion ambiguity and lower ill-posedness.  
Below is a concise summary of the five baseline methods.

\begin{itemize}
\item[-] \textbf{\cite{jin2018LearningExtract}}: The first learning-based blur decomposition method, which trains four separate sub-networks to progressively reconstruct the middle frame and the rest symmetric frames. Requiring to firstly recover the middle frame to serve as a reference for subsequent steps, it can only decompose blurry superimposition of videos comprising an odd number of frames. 

\item[-] \textbf{\cite{shedligeri2021UnifiedFramework}}: This method introduces a unified framework for recovering compressive videos from both coded aperture compressive temporal imaging and coded exposure imaging. Since it is originally designed to deal with grey-scale images, we extend it to RGB input by increasing corresponding network channels in the evaluation.

\item[-] \textbf{Animation from Blur} \citep{zhong2022AnimationBlur}: This method introduces motion guidance to facilitate the blur decomposition task and designs a unified framework supporting various input interfaces for motion guidance. For a fair comparison, we select the interface implemented with a motion prediction network, which could directly generate plausible motion guidance from a blurry measurement without additional motion annotation.

\item[-] \textbf{Slow Motion} \citep{jin2019LearningExtract}: Targeting to generate a sharp slow-motion video from a low-frame-rate blurry input, this algorithm consists of two main components: the DeblurNet estimating sharp keyframes, and the InterpNet predicting intermediate frames. By recursively calling the InterpNet, we up-convert the frame rate of the blurry test videos eight times to realize an equivalent performance comparison with our method.

\item[-] \textbf{BIN} \citep{shen2020BlurryVideo}: This algorithm also focuses on synthesizing high-frame-rate sharp videos from their low-frame-rate blurry counterparts. It incorporates a pyramid module for sharp intermediate frame estimation and an inter-pyramid recurrent module to exploit the temporal relationship. In the evaluation, we also recursively utilize this algorithm to achieve an $8 \times$ frame rate enhancement.
\end{itemize}

To ensure a fair comparison, we retrain all the methods except for \cite{jin2018LearningExtract} on the training set of GoPro. For \cite{jin2018LearningExtract}, we directly use the pre-trained weights on GoPro provided by the authors. During the evaluation, we use the sliding window strategy to select successive frames from every video in the GoPro test set and WAIC TSR dataset to generate the test samples. The length of the sliding window is set to 7 for \cite{jin2018LearningExtract} as it can only reconstruct videos with an odd number of frames. For other methods, the length of the sliding window is set to 8. Conforming to the physical process of imaging, there is no overlap between adjacent sliding windows.


\begin{figure}
\begin{center}
\includegraphics[width=0.92\linewidth]{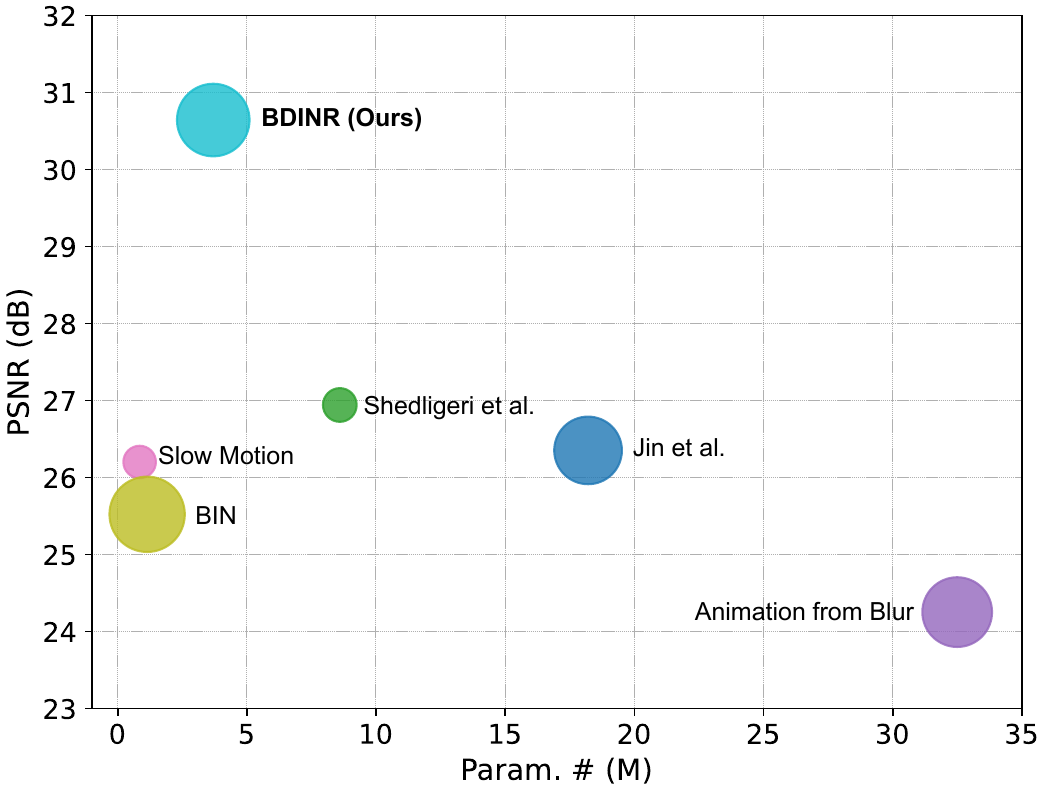}
\end{center}
\caption{\textbf{Params-PSNR-MACs comparsion with the comparative methods on GoPro.} The size of the bubbles represents the MACs index.} 
\label{fig:benchmark}
\end{figure}

\begin{figure*}
\begin{center}
  \includegraphics[width=0.9\linewidth]{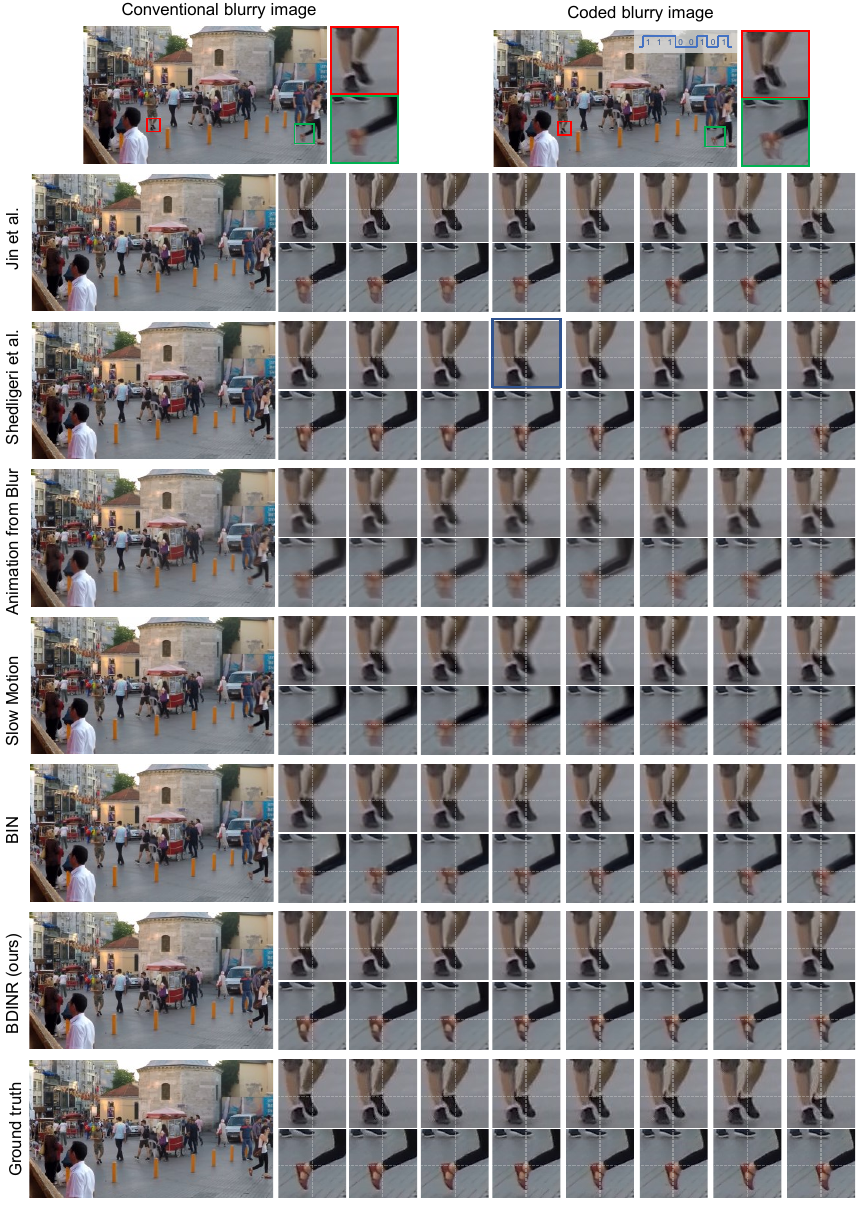}
\end{center}
\caption{\textbf{Qualitative comparison of BDINR with the baseline blur composition \citep{jin2018LearningExtract,shedligeri2021UnifiedFramework,zhong2022AnimationBlur} and blurry video interpolation  \citep{jin2019LearningExtract,shen2020BlurryVideo} methods on GoPro.} Note that blurry video interpolation methods take multiple adjacent frames as inputs, but we only demonstrate one of these frames here, similarly hereinafter. Please refer to the supplementary videos for a better visual perception of the temporal variation in the decomposed sequences.} 
\label{fig:simu_exp_gopro}
\end{figure*}

\begin{figure*}
\begin{center}
  \includegraphics[width=0.9\linewidth]{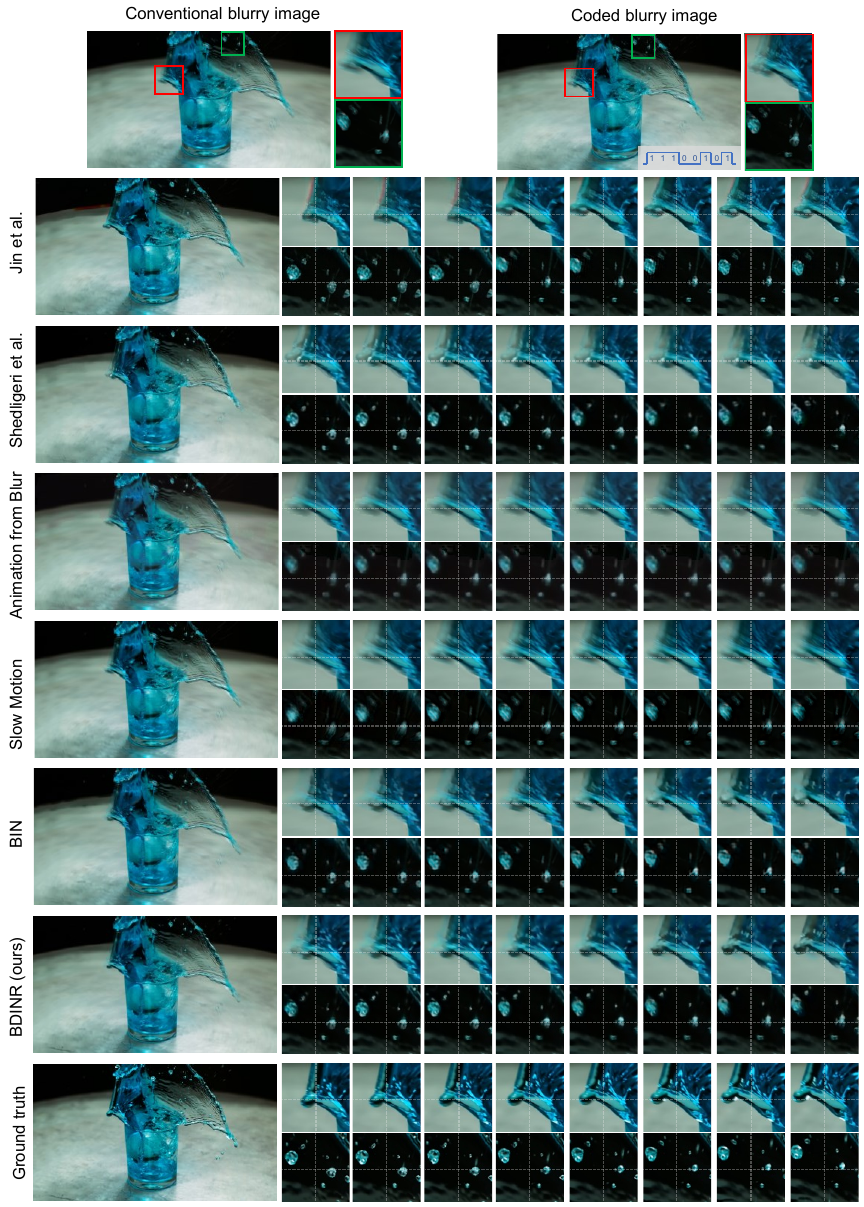}
\end{center}
\caption{\textbf{Qualitative comparison of BDINR with the comparative blur composition \citep{jin2018LearningExtract,shedligeri2021UnifiedFramework,zhong2022AnimationBlur} and blurry video interpolation methods \citep{jin2019LearningExtract,shen2020BlurryVideo} on WAIC TSR.} Please check out the supplementary videos for a better visual perception of the temporal variation in the blur decomposition results.}
\label{fig:simu_exp_waic}
\end{figure*}

\subsection{Results on synthetic Data.}  

We present the numerical results of the simulation experiments conducted on the GoPro and WAIC TSR datasets in Table~\ref{tab:simu_res}. Overall, our proposed method demonstrates superior performance, surpassing the second-ranked approach by a significant margin across all three metrics on both datasets. 
Additionally, Table~\ref{tab:simu_res} and Fig.~\ref{fig:benchmark} provide insights into the model size, MACs, and time required for extracting or interpolating a single frame for various algorithms. Our BDINR model features a lightweight architecture with only 3.7M parameters, thanks to its efficient INR-based video representation and self-recursive reconstruction strategy. While exhibiting moderate computational complexity and reconstruction speed compared to competitors, BDINR features higher flexibility in balancing the trade-off between computational burden and decomposition ratio during inference. A detailed exploration of this capability will be presented in Sec.~\ref{sec:analysis}.

We further demonstrate some results from BDINR and the competing methods for qualitative comparison in Fig.~\ref{fig:simu_exp_gopro} and Fig.~\ref{fig:simu_exp_waic}. As depicted in the figures, the proposed method successfully restores the details in the latent video frames while maintaining their correct ordering information. On the contrary, other blur decomposition methods \citep{jin2018LearningExtract,shedligeri2021UnifiedFramework,zhong2022AnimationBlur} either suffer from obvious blur artifacts or have difficulty figuring out the correct order among the retrieved video frames due to the ambiguity of the motion. These issues become even more severe in scenarios with complex object motions, as shown in Fig.~\ref{fig:simu_exp_waic}. In contrast, the blurry
video interpolation methods \citep{jin2019LearningExtract,shen2020BlurryVideo} face less challenge in retrieving the ordering information because they take multiple adjacent frames as input to mitigate the motion ambiguity. However, they still exhibit strong artifacts in the dynamic regions of the output videos.

\subsection{Data Capture and Results on Real Data}
\begin{figure}[h]
\begin{center}
  \includegraphics[width=0.9\linewidth]{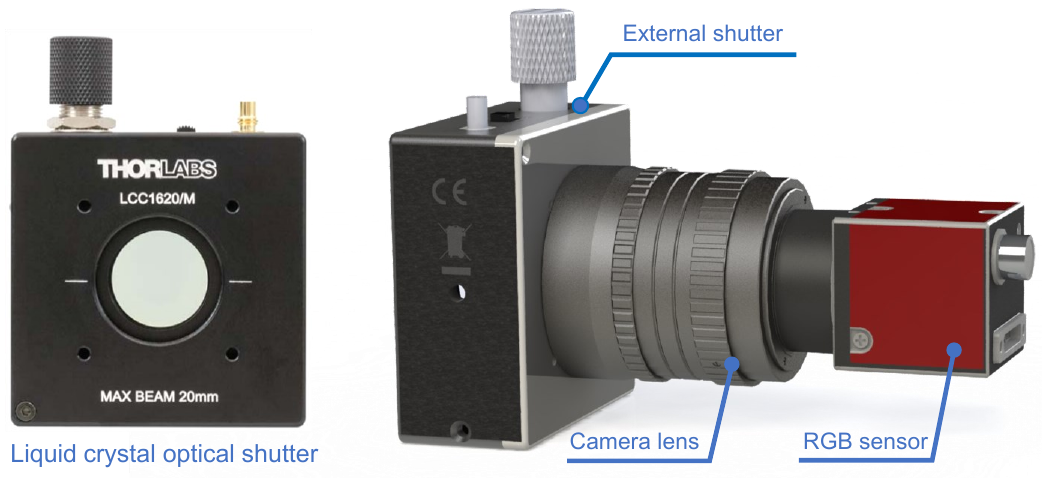}
\end{center}
\caption{\textbf{The prototype system for coded exposure photography.} A controllable liquid crystal optical shutter is mounted in front of the camera lens and synchronized with the RGB sensor to realize exposure encoding.} 
\label{fig:proto}
\end{figure}

\noindent\textbf{Prototype system.\quad} 
We build a prototype system of coded exposure photography to validate the effectiveness of the proposed method in practical settings. Generally, coded exposure photography can be directly realized using a camera that supports IEEE DCAM Trigger Mode 5. But here for high compatibility with most commercial cameras, we implement the system by introducing an extra external shutter synchronized by a micro-controller.
As shown in Fig~.\ref{fig:proto}, the system comprises a conventional RGB camera, a micro-controller, and an additional optical shutter. During acquisition, the micro-controller generates assigned binary voltage signals to control the open/close state of the optical shutter. Meanwhile, the micro-controller also synchronizes the camera with the shutter through a trigger signal.

\begin{figure*}[t]
\begin{center}
  \includegraphics[width=1\linewidth]{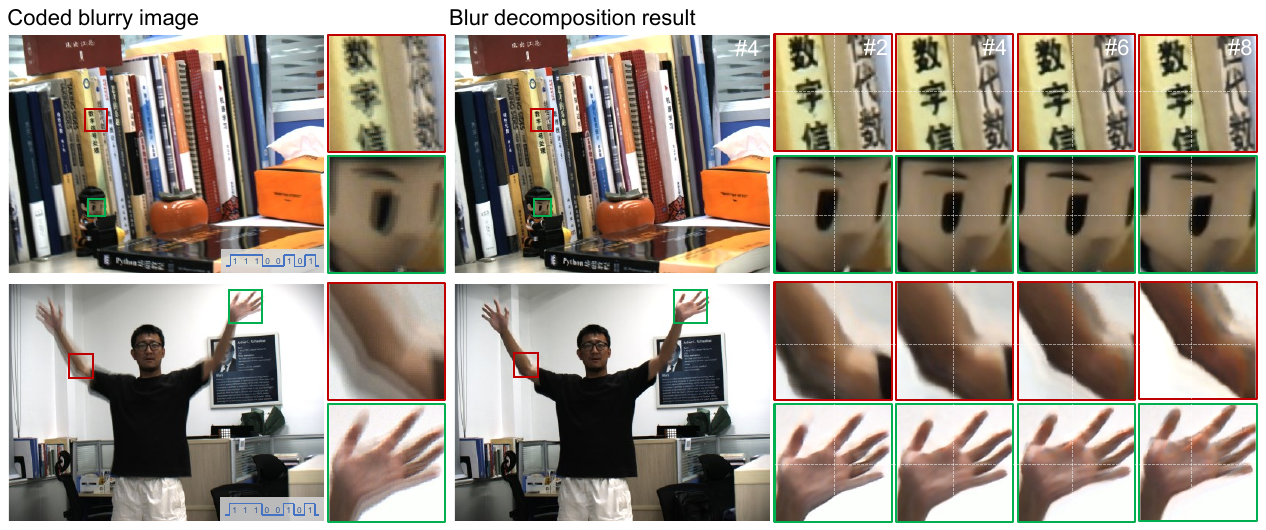}
\end{center}
\caption{\textbf{Blur decomposition results of the proposed framework on real-captured coded blurry images.} The upper row and the lower row demonstrate the blur decomposition results of different types of blurry images degraded by camera shake and object motion, respectively.} 
\label{fig:real_exp}
\end{figure*}

\vspace{1mm}
\noindent\textbf{Real-data results.\quad}
We captured coded blurry snapshots using the built prototype system and employed the pre-trained BDINR to extract the corresponding latent sharp video sequences. Here we examine the output of two kinds of typical blurs---caused by camera shake and object motion, with diverse forms of deterioration and even extensive spatial variation. The results, depicted in Fig.~\ref{fig:real_exp}, showcase the capability of our proposed method to successfully retrieve individual frames in the correct temporal arrangement for both types of blurs, demonstrating its versatility across a broad range of scenarios.

\begin{figure*}[!h]
\begin{center}
  \includegraphics[width=1\linewidth]{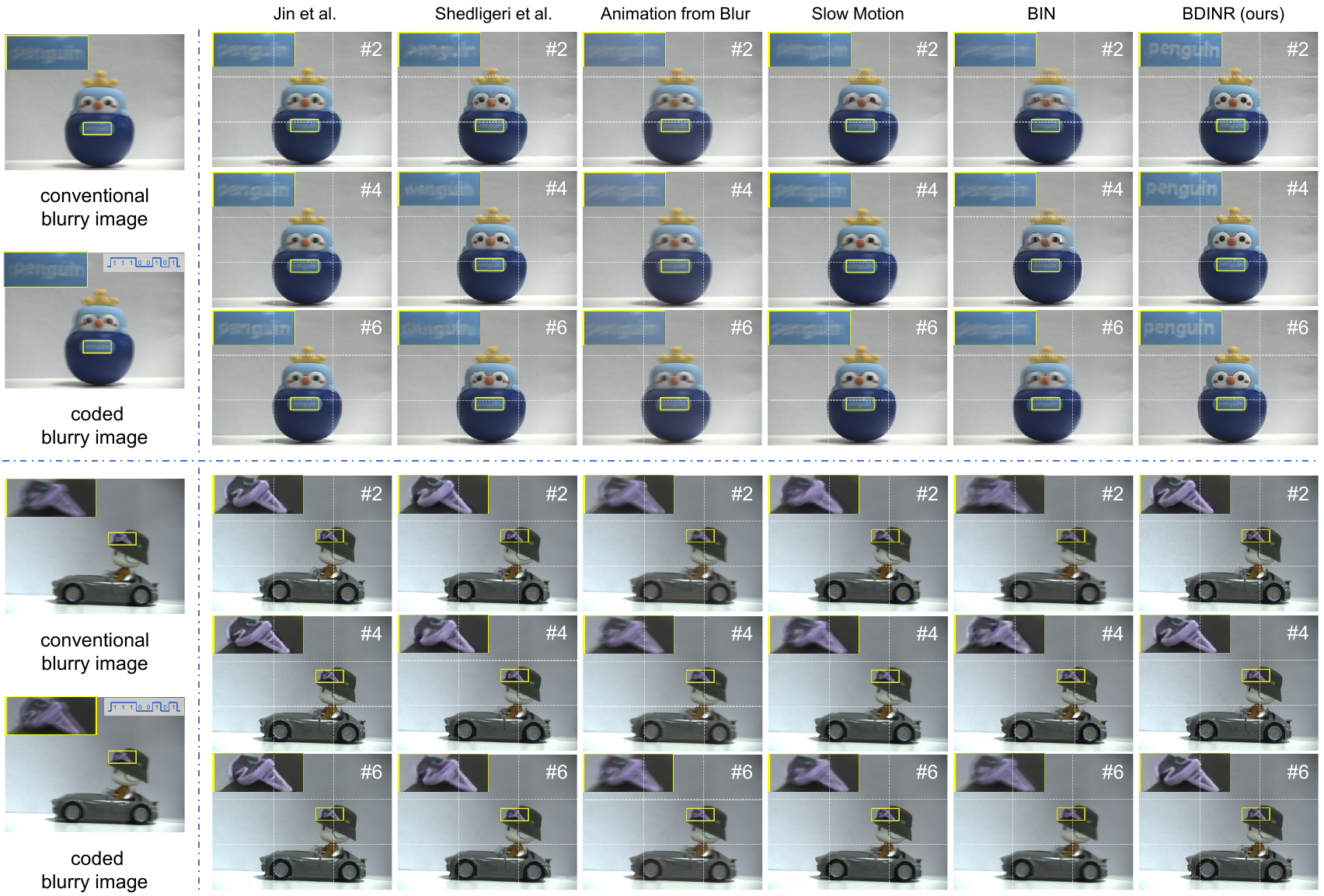}
\end{center}
\caption{\textbf{Qualitative comparison of BDINR with the comparative blur composition \citep{jin2018LearningExtract,shedligeri2021UnifiedFramework,zhong2022AnimationBlur} and blurry video interpolation methods \citep{jin2019LearningExtract,shen2020BlurryVideo} on real-world data.} Please check out the supplementary videos for a better visual perception.}
\label{fig:real_exp_cmp}
\end{figure*}

Furthermore, we conducted a comparative analysis of blur decomposition performance between BDINR and the competing methods using real-world data. While BDINR leverages coded exposure imaging, other methods mainly rely on conventional imaging. To ensure a fair comparison, repeatable scenes are necessary for acquiring twice with different exposure settings to yield similar video contents. Consequently, we conducted real-world experiments on controllable scenes, including a swinging toy penguin and a translating toy car, respectively. The results, depicted in Fig.~\ref{fig:real_exp_cmp}, illustrate that our proposed method adeptly restores coherent and sharp videos with fine details and much less artifacts.

\begin{figure}[t]
\begin{center}
  \includegraphics[width=0.95\linewidth]{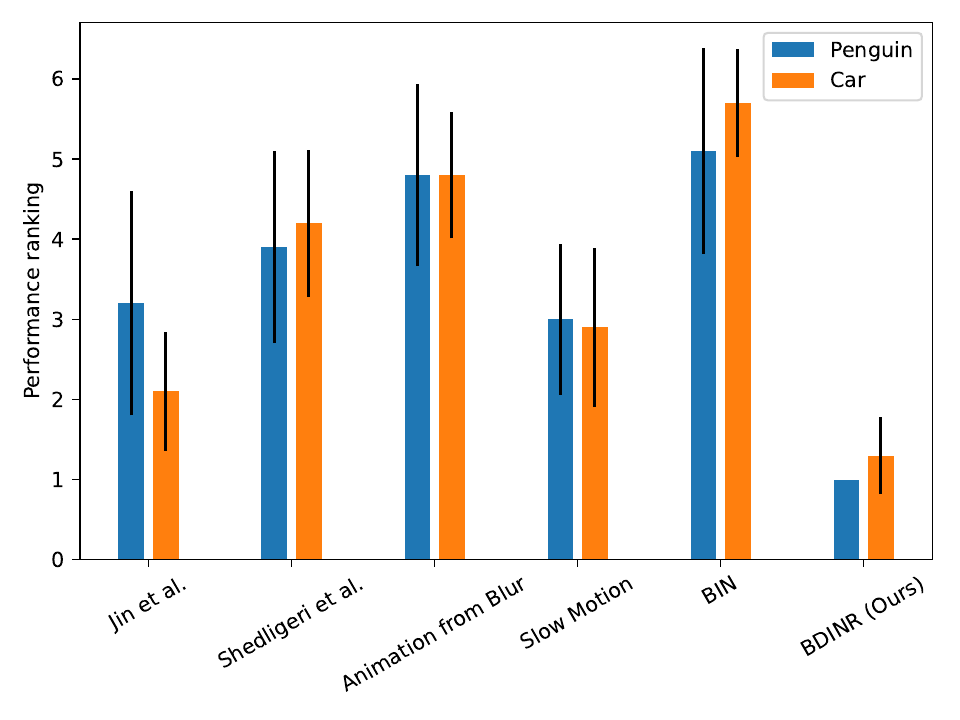}
\end{center}
\caption{\textbf{User study on different methods in real-data experiments.} Lower rankings indicate better performance.} 
\label{fig:real_exp_us}
\end{figure}

Given the absence of ground truth in real-world experiments, we quantitatively evaluated the reconstruction quality of various algorithms using state-of-the-art blind image quality assessment metrics, including MUSIQ \citep{ke2021MUSIQMultiscale} and DBCNN \citep{zhang2020BlindImage}. The results are summarized in Table \ref{tab:real_res}. Additionally, a user study was incorporated for subjective evaluation, with statistical analysis presented in Fig.~\ref{fig:real_exp_us}. In this study, ten individuals independently ranked the quality of the anonymized reconstruction results from different algorithms. Both the objective metrics and the subjective evaluation affirm that BDINR achieves superior performance to the competing methods, further substantiating its effectiveness in real-world scenarios.

\begin{table*}[t]
    \centering
    \caption{\textbf{Quantitative performance comparison with baseline blur decomposition and blurry video interpolation algorithms on real-world data.} \textbf{Bold} and \underline{underline} highlight the best and second-best scores, respectively; a higher score indicates superior performance.}
    \begin{center}
    \resizebox{1\textwidth}{!}{
    \begin{tabular}{llccccccc}
    \toprule
    & & Jin et al.& Shedligeri et al. & Animation from Blur  & Slow Motion & BIN & BDINR \\ 
    & & \cite{jin2018LearningExtract} & \cite{shedligeri2021UnifiedFramework} & \citep{zhong2022AnimationBlur} & \citep{jin2019LearningExtract} & \citep{shen2020BlurryVideo} & (Ours) \\ \midrule    
    \multirow{2}{*}{Penguin} & MUSIQ  & 42.22  & \underline{44.56}  & 39.53  & 43.25  & 39.40  & \textbf{49.77}  \\
    & DBCNN & \underline{35.91} & 33.97  & 32.14 & 31.40 & 32.39 & \textbf{36.99} \\ \midrule
    \multirow{2}{*}{Car} & MUSIQ & \textbf{43.82}  & 38.80  & 30.65  & 40.73  & 34.10  & \underline{43.43}  \\
    & DBCNN  & 30.46 & 28.76 & \underline{33.91}  & 30.50 & 27.91 & \textbf{33.93} \\
    \bottomrule
    \end{tabular}}
    \end{center}
    \label{tab:real_res}
\end{table*}

\begin{figure*}[h]
\begin{center}
  \includegraphics[width=1\linewidth]{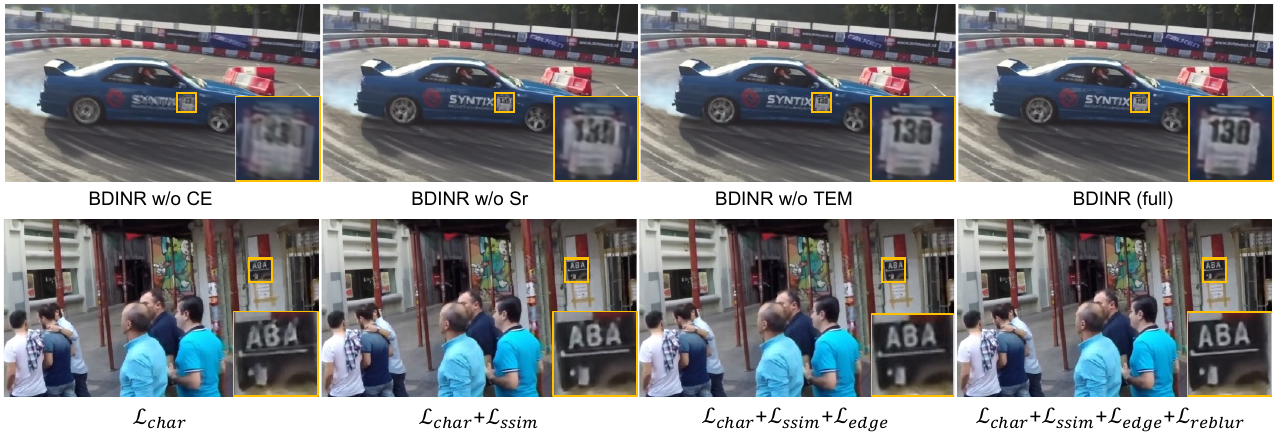}
\end{center}
\caption{\textbf{Visual results of blur decomposition regarding the ablation experiments.} The upper row corresponds to the ablation study on the network architecture. CE, Sr, and TEM refer to the coded exposure paradigm, self-recursive strategy, and temporal embedding module, respectively. The lower row corresponds to the ablation study on the loss function.} 
\label{fig:ablation_exp}
\end{figure*}

\subsection{Ablation Studies}
We conduct several ablation experiments to highlight the contributions of the key designs involved in the proposed framework.

\vspace{1mm}
\noindent\textbf{Coded exposure photography.\quad}
One of the main contributions of this work lies in introducing the coded exposure imaging technique for implicit embedding of the motion direction, targeting for tackling the motion direction ambiguity in blur decomposition. To demonstrate the advantage of this paradigm, we conduct an ablation study by removing the coded exposure strategy and retraining the network with the same settings as before. As evident from the results presented in Table~\ref{tab:ablation_arch} and Fig.~\ref{fig:ablation_exp}, the application of coded exposure photography distinctly enhances blur decomposition performance. Specifically, there is a significant increase of 4.52 dB on GoPro and 2.13 dB on WAIC TSR, underscoring the effectiveness of the coded exposure technique in mitigating motion direction ambiguity.

\begin{table}[h]  
\centering
\vspace{-3mm}
\caption{\textbf{Ablation study on the coded exposure photography and BDINR's architecture design.} \texttt{CE}, \texttt{Sr}, and \texttt{TEM} refer to coded exposure paradigm, self-recursive strategy, and temporal embedding module, respectively. In each case, the network is retrained from scratch on GoPro training set until convergence and then evaluated on GoPro test set and WAIC TSR dataset in terms of PSNR (dB)/SSIM. }
\label{tab:ablation_arch}
\centering
\resizebox{1\linewidth}{!}{%
\scriptsize
\begin{tabular}{c|cc}
\toprule
Variants & GoPro          & WAIC TSR \\ \midrule
BDINR w/o CE  & 26.12/0.8366 & 27.19/0.8634   \\
BDINR w/o Sr & 30.10/0.9045 & 28.00/0.8745  \\
BDINR w/o TEM  & 30.45/0.9125 & 28.38/0.8819\\ \midrule
BDINR (full)  & 30.64/0.9155 & 29.32/0.8805   \\  \bottomrule
\end{tabular}
}
\vspace{-3mm}
\end{table}

\vspace{1mm}
\noindent\textbf{Network architecture of BDINR.\quad}
Two key modules in the proposed blur decomposition algorithm are the self-recursive video INR module (INRV) and the incorporated temporal embedding module (TEM). The former sequentially extracts the latent sharp video frames from the coded blurry image, while the later  
efficiently models the temporal correlation and exploits cues of the embedded motion direction. 
We quantitatively validate the contributions of the self-recursive strategy and TEM by separately removing them from BDINR (i.e., BDINR w/o Sr and BDINR w/o TEM) and retraining the network from scratch on GoPro training set. The evaluation is performed on both GoPro test set and WAIC TSR dataset, and the results are summarized in Table~\ref{tab:ablation_arch}. The table demonstrates that the self-recursive strategy notably enhances the PSNR of BDINR by 0.54 dB and 1.32 dB on GoPro and WAIC TSR, respectively. Likewise, the TEM module contributes to improved PSNR values for BDINR, yielding enhancements of 0.19 dB and 0.94 dB on GoPro and WAIC TSR, respectively. Fig.~\ref{fig:ablation_exp} provides additional visual comparisons of various architecture configurations, showcasing the contributions of these network modules.

\vspace{1mm}
\noindent\textbf{Loss Function.\quad} 
We quantitatively assess the impact of each term in the loss function by incrementally incorporating individual components and retraining the network from scratch. The performance evaluation is conducted on the GoPro dataset, with results summarized in Table~\ref{tab:ablation_loss} and illustrated in Fig.~\ref{fig:ablation_exp}. Our analysis reveals that the inclusion of the SSIM loss and the Reblur loss leads to significant improvements in PSNR by 1.33 dB and 0.38 dB, respectively. Although the edge loss marginally enhances PSNR and SSIM scores, it contributes to enhancing the visual sharpness of the reconstructed results.

\begin{table}[h]
\centering
\caption{\textbf{Ablation study on the loss function design.} In each case, the network is retrained from scratch on GoPro until convergence and then evaluated on GoPro test set.}
\label{tab:ablation_loss}
\renewcommand{\arraystretch}{1.3}
\begin{tabular}{cccc|c}
\toprule
\multicolumn{3}{c}{Blur decomposition   loss} & \multirow{2.1}{*}{Reblur loss} & \multirow{2.1}{*}{PSNR / SSIM}  \\\cmidrule(lr){1-3}
$\mathcal{L}_{char}$ & $\mathcal{L}_{ssim}$ & $\mathcal{L}_{edge}$ & & \\ \midrule
\checkmark & & & & 28.89 / 0.8849  \\
\checkmark & \checkmark &  & & 30.22 / 0.9078  \\
\checkmark & \checkmark & \checkmark &  & 30.26 / 0.9091  \\
\checkmark & \checkmark & \checkmark & \checkmark & 30.64 / 0.9155  \\
\bottomrule
\end{tabular}
\vspace{-3mm}
\end{table}

\subsection{Discussions and Analysis}
\label{sec:analysis}
This subsection firstly investigates the influence of coded-exposure sequences on the performance of the proposed blur decomposition framework. It then highlights BDINR's flexibility in selective reconstruction of latent frames. Afterward, we summary the limitations and prospects of the current implementation.

\begin{table*}[]
\caption{\textbf{The influence of the length and duty ratio of the coded-exposure sequence on BDINR's performance.} In each case, the network is retrained from scratch on GoPro training set until convergence and then evaluated on GoPro test set.}
\label{analysis}
\centering
\resizebox{1\linewidth}{!}{%
\renewcommand{\arraystretch}{1.2}
\begin{tabular}{c|ccccccccccccc}
\toprule
\multirow{2.2}{*}{~~~~Performance~~~~} & &\multicolumn{5}{c}{\quad\quad\quad Length of sequence (\# of bits)\quad\quad\quad }& & \multicolumn{5}{c}{\quad\quad\quad\quad Duty ratio\quad\quad } \\ \cmidrule(lr{0pt}){3-7} \cmidrule(lr{0pt}){9-14}
&& 4 & 6 & 8 & 10 & 12 && 3/8 & 4/8 & 5/8 & 6/8 & 7/8 & 8/8\\ \midrule
PSNR (dB)  && 34.60  & 30.98  & 30.64  & 25.72  & 25.21  && 26.02  & 27.17  & 30.64  & 30.24  & 30.25 & 26.12\\
SSIM  && 0.9615 & 0.9196 & 0.9155 & 0.8090 & 0.7898 && 0.8338 & 0.8481 & 0.9155 & 0.9062 & 0.9082 & 0.8366 \\
LPIPS && 0.0527 & 0.1028 & 0.1202 & 0.2197 & 0.2575 && 0.1817 & 0.1785 & 0.1202 & 0.1325 & 0.1273 & 0.1810 \\ \bottomrule
\end{tabular}
}
\end{table*}

\begin{figure*}[h]
\begin{center}
  \includegraphics[width=1\linewidth]{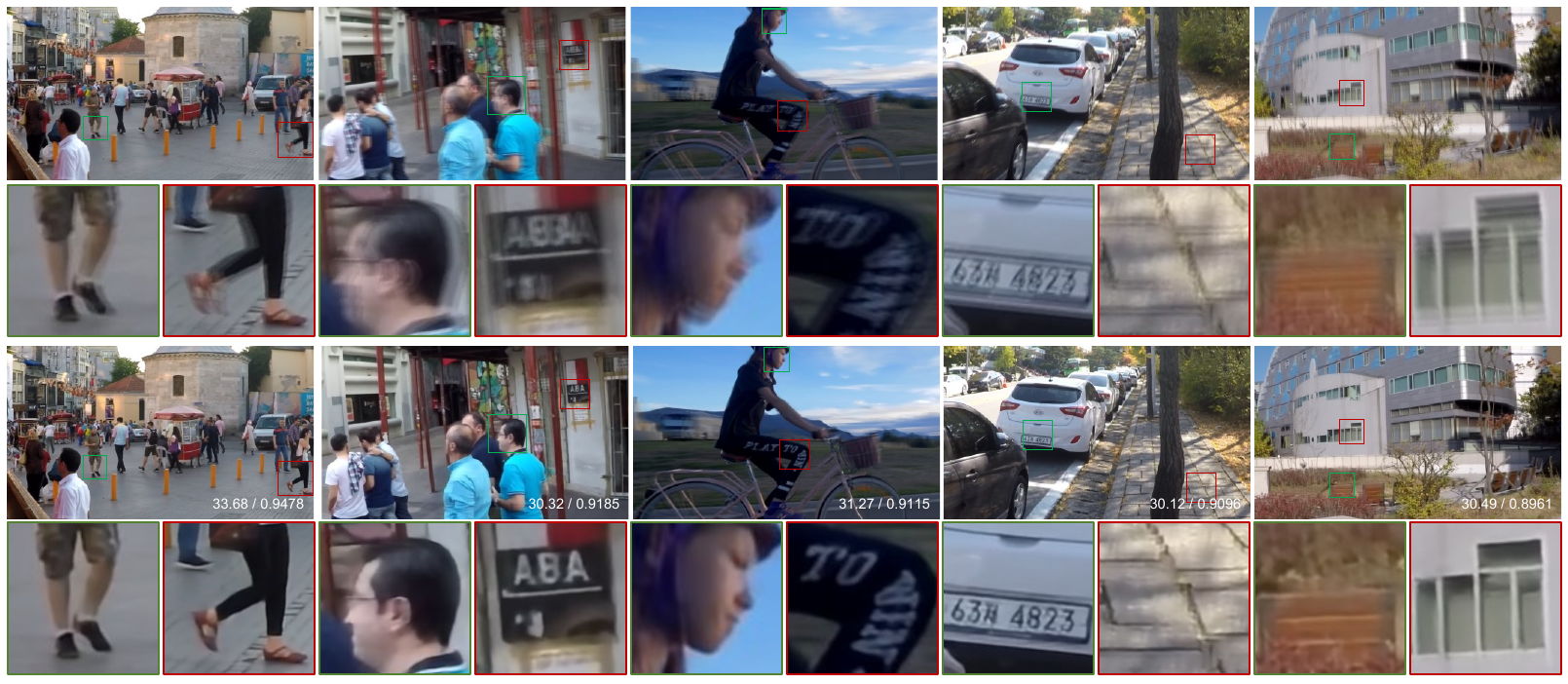}
\end{center}
\caption{\textbf{Motion deblurring examples with BDINR}. Benefiting from the selective extraction capability, BDINR can be regarded as a motion deblurring algorithm by extracting only the middle frame during inference. The deblurring performance in terms of PSNR (dB) / SSIM is labelled on the bottom-right corner of the deblurred images.} 
\label{fig:deblur}
\end{figure*}

\vspace{1mm}
\noindent\textbf{Encoding sequences of coded exposure.\quad}
The length and duty ratio of the temporal encoding sequence are two key hyper-parameters in the implementation of coded exposure photography. We conduct a series of experiments to investigate their influences on BDINR's performance and present the results in Table~\ref{analysis}.

Assuming each bit in the encoding sequence corresponds to a constant duration during image acquisition, a longer encoding sequence will result in a longer exposure period. In this process, more scenario information will be embedded into the captured coded blurry snapshot, and the blur decomposition network needs to extract more sharp frames during the post-processing accordingly. Briefly speaking, a longer encoding sequence indicates a higher compression ratio and a heavier burden for the blur decomposition algorithm. The results are consistent with the intuition, as shown in Table~\ref{analysis}, from which one can see that as the encoding length increases, the performance of BDINR drops consistently. Furthermore, when the information entropy of the coded blurry image surpasses the maximum representation capacity of the video INR module incorporated in BDINR, the blur decomposition performance will suffer from a significant decline.

The duty ratio of the encoding sequence is defined as the proportion of `1's in the encoding sequence, which physically determines the light throughput of the coded exposure imaging system and tightly correlates with the signal-to-noise ratio (SNR) of the coded measurement. The results in Table~\ref{analysis} unveil two notable trends: (i) BDINR's performance decreases rapidly as the duty ratio drops below $\frac{5}{8}$ for 8-bit exposure encoding sequences; (ii) a greater duty ratio may also cause a minor decrease in performance due to excessive information coupling. Additionally, Table~\ref{analysis} encompasses the results for the duty ratio of $\frac{8}{8}$, indicating the absence of coded exposure technique utilization. In this scenario, the PSNR stands at 26.12dB, approximately 4dB lower compared to employing the coded exposure technique. This substantial contrast underscores the effectiveness of coded exposure in enhancing blur decomposition performance.

\vspace{1mm}
\noindent\textbf{Selective extraction of latent frames.\quad}
BDINR offers remarkable flexibility by enabling selective reconstruction of latent frames rather than extracting all of them during inference. This capability is facilitated by its INR-based video representation and disentangled spatial-temporal context inputs. Such flexibility allows BDINR to retrieve desired frames given their index as inputs, providing a strategic trade-off between computational burden and decomposition ratio. For instance, when only one frame is needed, BDINR seamlessly transitions into a motion-deblurring network. As illustrated in Fig.~\ref{fig:deblur}, without requiring any retraining or fine-tuning, BDINR demonstrates impressive deblurring performance across various motion-blurred images, with significantly reduced time consumption compared to decomposing all latent frames.

However, it's important to note that this selective extraction capacity is not compatible with the self-recursive reconstruction strategy, which relies on adjacent frame information during both training and inference. In other words, the flexibility of selective extraction comes at the expense of discarding the self-recursive strategy, leading to a slight decline in performance.

\vspace{1mm}
\noindent\textbf{Limitations and prospects.\quad} 
The proposed framework primarily targets the blur decomposition of individual blurry images. Consequently, slight temporal inconsistencies may arise between the sharp images decomposed from adjacent blurry measurements when applied to blurry videos. Currently, this issue can be alleviated by averaging transitional frames in the output video to ensure smooth transitions. Expanding the proposed method to address blurry video temporal super-resolution by leveraging inter-frame correlation presents a feasible and promising avenue for future exploration. In this case, the motion direction ambiguity will be further alleviated, and the design of encoding sequences for coded exposure photography can focus more on preventing high-frequency loss and facilitating the decomposition network to pursue better reconstruction quality.

Further research avenues also include the joint optimization of the exposure encoding sequence and the blur decomposition network in an end-to-end fashion to achieve comprehensive performance enhancement. Additionally, extending the application of coded exposure photography to encompass tasks such as space-time super-resolution of blurry videos \citep{Geng_2022_CVPR} and 3D scene reconstruction from motion-blurred images \citep{qiu2019WorldBlur} offers compelling opportunities for further exploration.
Moreover, investigating the utilization of coded illumination instead of coded exposure to achieve more precise multi-level exposure control in specialized applications such as microscopy warrants thorough investigation.

Besides, we will build a low-cost high-speed coded shutter and further trim the model to reduce the computing resources, and deploy it as an add-on to the commercial vision platforms such as smartphone cameras.

\section{Conclusion}
\label{sec:conclusion}

We revisit the coded exposure photography to achieve lightweight high-speed photography at high resolution and low bandwidth, via developing a novel blur decomposition framework incorporating the forward imaging model of coded exposure and implicit neural representation of natural videos. The framework tactfully tackles the challenge of motion direction ambiguity by implicitly embedding the motion direction cues into the coded blurry image during data acquisition. It also incorporates a learnable video INR empowered self-recursive neural network to exploit the embedded cues indicating motion direction for high-quality sharp video sequence retrieval. 

Though the framework is proposed for blur decomposition of single blurry images, it can also be flexibly extended to blurry video temporal super-resolution and motion deblurring without any modification or retraining. Compared with existing blur decomposition approaches, the proposed framework has advantages in low system complexity, small model size, and high application flexibility. We believe that it will open a promising avenue for low-bandwidth, low-cost, high-speed imaging and shed new light on applications of mobile vision systems, including video surveillance, video assistant referee, auto-driving, inspection, etc.

\section*{Supplementary Information}
The supplementary material contains three videos that demonstrate the blur decomposition results of the proposed framework and its comparison with the competing methods.

\section*{Acknowledgment}
This work was supported by the Ministry of Science and Technology of the People’s Republic of China [grant number 2020AAA0108202] and the National Natural Science Foundation of China [grant numbers 61931012, 62088102].

\section*{Data Availability}
The datasets generated during and/or analysed during the current study are available from the corresponding author on reasonable request.

\section*{Conflict of Interest}
The authors have no relevant financial or non-financial interests to disclose.

\bibliographystyle{spbasic}
\bibliography{ref}

\end{sloppypar}
\end{document}